\documentclass[runningheads]{llncs}
\usepackage{graphicx}
\usepackage[utf8]{inputenc} 
\usepackage[T1]{fontenc}    
\usepackage{hyperref}       
\usepackage{url}            
\usepackage{booktabs}       
\usepackage{amsfonts}       
\usepackage{nicefrac}       
\usepackage{microtype}      
\usepackage{xcolor}         

\usepackage{rotating}
\usepackage{amsmath,amsfonts,amscd,amssymb}
\usepackage{mathtools}
\usepackage{dsfont}
\usepackage{tikz}
\usepackage[mathscr]{euscript}
\usepackage[noend]{algorithmic}
\usepackage{algorithm}
\usepackage{setspace}
\usepackage{lineno}
\usetikzlibrary{decorations.pathreplacing}
\definecolor{color1bg}{HTML}{f73d28}
\definecolor{color2bg}{HTML}{FA8072}
\definecolor{bblue}{HTML}{00BFFF}
\definecolor{bblue2}{HTML}{00ffff}

\usetikzlibrary{arrows,calc}
\tikzset{
>=stealth',
help lines/.style={dashed, thick},
axis/.style={<->},
important line/.style={thick},
connection/.style={thick, dotted},
}

\usetikzlibrary{shadows}
\usetikzlibrary{backgrounds}
\tikzset{
diagonal fill/.style 2 args={fill=#2, path picture={
		\fill[#1, sharp corners] (path picture bounding box.south west) -|
		(path picture bounding box.north east) -- cycle;}},
reversed diagonal fill/.style 2 args={fill=#2, path picture={
		\fill[#1, sharp corners] (path picture bounding box.north west) |- 
		(path picture bounding box.south east) -- cycle;}}
}
\usetikzlibrary{arrows.meta}

\begin{document}
\title{An Algorithm to Train Unrestricted Sequential Discrete Morphological Neural Networks\thanks{Corresponding author: D. Marcondes (\email{dmarcondes@ime.usp.br}). D. Marcondes was funded by grants \#22/06211-2 and \#23/00256-7, São Paulo Research Foundation (FAPESP), and J. Barrera was funded by grants \#14/50937-1 and \#2020/06950-4, São Paulo Research Foundation (FAPESP).}}
\titlerunning{Unrestricted Sequential Discrete Morphological Neural Networks}
%
\author{Diego Marcondes\inst{1,2}\orcidID{0000-0002-6087-4821} \and
	Mariana Feldman\inst{1}\orcidID{0009-0002-1961-9172} \and
	Junior Barrera\inst{1}\orcidID{0000-0003-0439-0475}}
\authorrunning{D. Marcondes, M. Feldman and J. Barrera}
%
\institute{Department of Computer Science, Institute of Mathematics and Statistics, University of São Paulo, São Paulo, Brazil  \and
	Department of Electrical and Computer Engineering, Texas A\&M University, College Station, USA}
\maketitle              
\begin{abstract}
	There have been attempts to insert mathematical morphology (MM) operators into convolutional neural networks (CNN), and the most successful endeavor to date has been the morphological neural networks (MNN). Although MNN have performed better than CNN in solving some problems, they inherit their black-box nature. Furthermore, in the case of binary images, they are approximations that loose the Boolean lattice structure of MM operators and, thus, it is not possible to represent a specific class of W-operators with desired properties. In a recent work, we proposed the Discrete Morphological Neural Networks (DMNN) for binary image transformation to represent specific classes of W-operators and estimate them via machine learning. We also proposed a stochastic lattice descent algorithm (SLDA) to learn the parameters of Canonical Discrete Morphological Neural Networks (CDMNN), whose architecture is composed only of operators that can be decomposed as the supremum, infimum, and complement of erosions and dilations. In this paper, we propose an algorithm to learn unrestricted sequential DMNN, whose architecture is given by the composition of general W-operators. We illustrate the algorithm in a practical example.
	
\end{abstract}
\section{Introduction}

Mathematical morphology is a theory of lattice mappings which can be employed to design nonlinear mappings, the morphological operators, for image processing and computer vision. It originated in the 60s, and its theoretical basis was developed in the 70s and 80s \cite{matheron1974random,serra1982image,serra1988image}. From the established theory followed the proposal of many families of operators, which identify or transform geometrical and topological properties of images. Their heuristic combination permits to design methods for image analysis. We refer to \cite{dougherty2003hands} for more details on practical methods and implementations of the design of mathematical morphology operators.

Since combining basic morphological operators to form a complex image processing pipeline is not a trivial task, a natural idea is to develop methods to automatically design morphological operators based on machine learning techniques \cite{barrera2000automatic}, what has been extensively done in the literature with great success on solving specific problems. The problems addressed by mathematical morphology concern mainly the processing of binary and gray-scale images, and the learning methods are primarily based on discrete combinatorial algorithms. More details about mathematical morphology in the context of machine learning may be found in \cite{barrera2022mathematical,hirata2021machine}.

Mathematical morphology methods have also been studied in connection with neural networks. The first papers about morphological neural networks (MNN), such as \cite{davidson1992simulated,davidson1993morphology,davidson1990theory,ritter1996introduction}, proposed neural network architectures in which the operation performed by each neuron is either an erosion or a dilation. MNN usually have the general structure of neural networks, and their specificity is on the fact that the layers realize morphological operations. Many MNN architectures and training algorithms have been proposed for classification and image processing problems \cite{araujo2017morphological,dimitriadis2021advances,grana2001some,mondal2019morphological2,sussner2009constructive}. Special classes of MNN such as morphological/rank neural networks \cite{pessoa1996morphological,pessoa2000neural}, the dendrite MNN \cite{arce2018differential,ritter2003morphological,sossa2014efficient}; and the modular MNN \cite{araujo2006modular,de2000designing} have been proposed.  We refer to \cite{monteiro2008brief} and the references therein for a review of the early learning methods based on MNN.

More recently, mathematical morphology methods have been studied in connection with deep neural networks, by either combining convolutional neural networks (CNN) with a morphological operator \cite{julca2017image}, or by replacing the convolution operations of CNN with basic morphological operators, such as erosions and dilations \cite{franchi2020deep,groenendijk2022morphpool,mondal2019morphological}.  Although it has been seen empirically that convolutional layers could be replaced by morphological layers \cite{franchi2020deep}, and MNN have shown a better performance than CNN in some tasks \cite{hu2022learning}, they are not more interpretable than an ordinary CNN. In this context, the interpretability is related to the notion of hypothesis space, which is a key component of machine learning theory.

In \cite{marcondes2023discrete} we proposed the Discrete Morphological Neural Networks (DMNN) for binary image transformation to represent translation invariant and locally defined operators, i.e., W-operators, and estimate them via machine learning. A DMNN architecture is represented by a Morphological Computational Graph that combines operators via composition, infimum, and supremum to build more complex operators. In \cite{marcondes2023discrete} we proposed a stochastic lattice descent algorithm (SLDA) to train the parameters of Canonical Discrete Morphological Neural Networks (CDMNN) based on a sample of input and output images under the usual machine learning approach. The architecture of a CDMNN is composed only of canonical operations (supremum, infimum, or complement) and operators that can be decomposed as the supremum, infimum, and complement of erosions and dilations with the same structural element. The DMNN is a true mathematical morphology method since it retains the control over the design and the interpretability of results intrinsic to classical mathematical morphology methods, which is a relevant advantage over CNN.

In this paper, we propose an algorithm to train unrestricted sequential DMNN (USDMNN), whose architecture is given by the composition of general W-operators. The algorithm considers the representation of a W-operator by its characteristic Boolean function, which is learned via a SLDA in the Boolean lattice of functions. This SLDA differs from that of \cite{marcondes2023discrete} which minimizes an empirical error in a lattice of sets and intervals that are the structural elements and intervals representing operators such as erosions, dilations, openings, closings and sup-generating. 

Unlike CDMNN, which can be designed from prior information to represent a constrained class of operators, USDMNN are subject to overfitting the data, since it can represent a great class of operators, that may fit the data, but not generalize to new examples. In order to address this issue, we control the complexity of the architecture by selecting from data the window of each W-operator in the sequence (layer). By restricting the window, we create equivalence classes on the characteristic function domain that constrain the class of operators that each layer can represent, decreasing the overall complexity of the operators represented by the architecture and mitigating the risk of overfitting. We propose a SLDA to select the windows by minimizing a validation error in a lattice of sets within the usual model selection framework in machine learning.

We note that both the CDMNN and the USDMNN are fully transparent, the properties of the operators represented by the trained architectures are fully known, and their results can be interpreted. The advantage of USDMNN is that they are not as dependent on prior information as the CDMNN, which require a careful design of the architecture to represent a \textit{simple} class of operators with properties necessary to solve the practical problem, as was illustrated in the empirical application in \cite{marcondes2023discrete}. When there is no prior information about the problem at hand, an USDMNN may be preferable.

In Section \ref{SecPreliminaries}, we present some notations and definitions, and in Section \ref{SecModel} we formally define the USDMNN that is a particular example of the DMNN proposed by \cite{marcondes2023discrete}. In Section \ref{SecTrain}, we present the SLDAs to learn the windows and the characteristic functions of the W-operators in a USDMNN, and in Section \ref{SecToyExample} we apply the USDMNN to the same dataset used in \cite{marcondes2023discrete} in order to compare the results. In Section \ref{SecPerspectives}, we present the next steps of this research.

\section{W-operators}
\label{SecPreliminaries}

Let $E = \mathbb{Z}^{2}$ and denote by $\mathcal{P}(E)$ the collection of all subsets of $E$. Denote by $+$ the vector addition operation. We denote the zero element of $E$ by $o$. A \textit{set operator} is any mapping defined from $\mathcal{P}(E)$ into itself. We denote by $\Psi$ the collection of all the operators from $\mathcal{P}(E)$ to $\mathcal{P}(E)$. Denote by $\iota$ the identity set operator: $\iota(X) = X, X \in \mathcal{P}(E)$.

For any $h \in E$ and $X \in \mathcal{P}(E)$, the set $X + h \coloneqq \{x \in E: x - h \in X\}$ is called the translation of $X$ by $h$. We may also denote this set by $X_{h}$. A set operator $\psi$ is called \textit{translation invariant} (t.i) if, and only if, $\forall h \in E$, $\psi(X + h) = \psi(X) + h$ for $X \in \mathcal{P}(E)$. 

Let $W$ be a finite subset of $E$. A set operator $\psi$ is called \textit{locally defined within a window} $W$ if, and only if, $\forall h \in E$, $h \in \psi(X) \iff h \in \psi(X \cap W_{h})$ for $X \in \mathcal{P}(E)$.	The collection $\Psi_{W}$ of t.i. operators locally defined within a window $W \in \mathcal{P}(E)$ inherits the complete lattice structure of $(\mathcal{P}(E),\subseteq)$ by setting, $\forall \psi_{1},\psi_{2} \in \Psi_{W}$,
\begin{linenomath}
	\begin{align}
		\label{partial_Psi}
		\psi_{1} \leq \psi_{2} \iff \psi_{1}(X) \subseteq \psi_{2}(X), \forall X \in \mathcal{P}(E).
	\end{align}
\end{linenomath}
Define by $\Omega = \cup_{W \in \mathcal{P}(E), |W| < \infty} \Psi_{W}$ the collection of all operators from $\mathcal{P}(E)$ to $\mathcal{P}(E)$ that are t.i. and locally defined within some finite window $W \in \mathcal{P}(E)$. The elements of $\Omega$ are called $W$\textit{-operators}.

Any W-operator can be uniquely determined by its kernel, its basis or its characteristic function (see \cite{barrera1996set} for more details). In special, denote by $\mathfrak{B}_{W} \coloneqq \{0,1\}^{\mathcal{P}(W)}$ the set of all Boolean functions on $\mathcal{P}(W)$ and consider the mapping $T$ between $\Psi_{W}$ and $\mathfrak{B}_{W}$ defined by
\begin{linenomath}
	\begin{align}
		T(\psi)(X) = \begin{cases}
			1, & \text{ if } o \in \psi(X)\\
			0, & \text{ otherwise}.
		\end{cases} & & \psi \in \Psi_{W}, X \in \mathcal{P}(W).
	\end{align}
\end{linenomath}
The mapping $T$ constitutes a lattice isomorphism between the complete lattices $(\Psi_{W},\leq)$ and $(\mathfrak{B}_{W},\leq)$, and its inverse $T^{-1}$ is defined by $T^{-1}(f)(X) = \left\{x \in E: f(X_{-x} \cap W) = 1\right\}$ for $f \in \mathfrak{B}_{W}$ and $X \in \mathcal{P}(E)$. We denote by $f_{\psi} \coloneqq T(\psi)$ the characteristic function of $\psi$.

\section{Unrestricted Sequential Discrete Morphological Neural Networks}
\label{SecModel} 

In this section, we formally define the USDMNN. In Section \ref{Sec31}, we define the morphological computational graph of USDMNN. This is a concept introduced in \cite{marcondes2023discrete} that is related to their architecture, defined in Section \ref{Sec32}. In Section \ref{Sec33}, we propose a representation of USDMNN by the windows and characteristic functions of the operators in their layers.

\subsection{Sequential Morphological Computational graph}
\label{Sec31}

Let $\mathcal{G} = (\mathcal{V},\mathcal{E},\mathcal{C})$ be a \textit{computational graph}, in which $\mathcal{V}$ is a general set of vertices, $\mathcal{E} \subset \{(\mathfrak{v}_{1},\mathfrak{v}_{2}) \in \mathcal{V} \times \mathcal{V}: \mathfrak{v}_{1} \neq \mathfrak{v}_{2}\}$ is a set of directed edges, and $\mathcal{C}: \mathcal{V} \to \Omega \cup \{\vee,\wedge\}$ is a mapping that associates each vertex $\mathfrak{v} \in \mathcal{V}$ to a \textit{computation} given by either applying a t.i. and locally defined operator $\psi \in \Omega$ or one of the two basic operations $\{\vee,\wedge\}$.

Denoting $\mathcal{V} = \left\{\mathfrak{v}_{i},\mathfrak{v}_{1},\dots,\mathfrak{v}_{n},\mathfrak{v}_{o}\right\}$ for a $n \geq 1$, $\mathcal{G}$ is a sequential morphological computational graph (MCG) if
\begin{linenomath}
	\begin{equation}
		\mathcal{E} = \left\{(\mathfrak{v}_{i},\mathfrak{v}_{1}),(\mathfrak{v}_{n},\mathfrak{v}_{o})\right\} \bigcup \left\{(\mathfrak{v}_{j},\mathfrak{v}_{j+1}): j = 1,\dots,n-1\right\},
	\end{equation}
\end{linenomath}
with $\mathcal{C}(\mathfrak{v}_{i}) = \mathcal{C}(\mathfrak{v}_{o}) = \iota$ and $\mathcal{C}(\mathfrak{v}_{j}) \in \Omega, \ j = 1,\dots,n$. This computational graph satisfies the axioms of MCG (see \cite{marcondes2023discrete} for more details).

In a sequential MCG, the computation of a vertex $\mathfrak{v}_{j}$ in $\mathcal{G}$ receives as input the output of the computation of the previous vertex $\mathfrak{v}_{j-1}$,  and the output of its computation will be used as the input of the computation of the vertex $\mathfrak{v}_{j+1}$. We assume there is an input vertex $\mathfrak{v}_{i}$, and an output vertex $\mathfrak{v}_{o}$, that store the input, which is an element $X \in \mathcal{P}(E)$, and output of the computational graph, respectively, by applying the identity operator. Furthermore, each vertex computes an operator in $\Omega$ and there are no vertices computing supremum or infimum operations.

Denote by $\psi_{\mathcal{G}}(X)$ the output of vertex $\mathfrak{v}_{o}$ when the input of vertex $\mathfrak{v}_{i}$ is $X \in \mathcal{P}(E)$ and let $\psi_{\mathcal{G}}: \mathcal{P}(E) \to \mathcal{P}(E)$ be the set operator generated by MCG $\mathcal{G}$. The operator $\psi_{\mathcal{G}}$ is actually t.i. and locally defined within a window $W_{\mathcal{G}}$ (cf. Proposition 5.1 in \cite{marcondes2023discrete}). We define the sequential Discrete Morphological Neural Network represented by $\mathcal{G}$ as the translation invariant and locally defined set operator $\psi_{\mathcal{G}}$.	

\subsection{Sequential Discrete Morphological Neural Networks Architecture}
\label{Sec32}

A triple $\mathcal{A} = (\mathcal{V},\mathcal{E},\mathcal{F})$, in which $\mathcal{F} \subseteq \Omega^{\mathcal{V}}$, is a sequential Discrete Morphological Neural Network (SDMNN) architecture if $(\mathcal{V},\mathcal{E},\mathcal{C})$ is a sequential MCG for all $\mathcal{C} \in \mathcal{F}$. A SDMNN architecture is a collection of sequential MCG with the same graph $(\mathcal{V},\mathcal{E})$ and computation map $\mathcal{C}$ in $\mathcal{F}$. Since a SDMNN architecture represents a collection of MCG, it actually represents a collection of t.i. and locally defined set operators that can be represented as the composition of W-operators. 

For an architecture $\mathcal{A} = (\mathcal{V},\mathcal{E},\mathcal{F})$, let $\mathbb{G}(\mathcal{A}) = \left\{\mathcal{G} = (\mathcal{V},\mathcal{E},\mathcal{C}): \mathcal{C} \in \mathcal{F}\right\}$ be the collection of MCG generated by $\mathcal{A}$. We say that $\mathcal{G} \in \mathbb{G}(\mathcal{A})$ is a realization of architecture $\mathcal{A}$ and we define $\mathcal{H}(\mathcal{A}) = \left\{\psi \in \Omega: \psi = \psi_{\mathcal{G}}, \mathcal{G} \in \mathbb{G}(\mathcal{A})\right\}$ as the collection of t.i. and locally defined set operators that can be realized by $\mathcal{A}$.

A SDMNN is said unrestricted if its interior vertices $\mathfrak{v}_{j}$ can compute any W-operator locally defined within a $W_{j}$, so it holds $\mathcal{F} = \{\iota\} \times \Psi_{W_{1}} \times \cdots \times \Psi_{W_{n}} \times \{\iota\}$. We denote by $n = |\mathcal{V}| - 2$ the depth of an unrestricted sequential DMNN (USDMNN) and by $|W_{j}|$ the width of the layer represented by vertex $\mathfrak{v}_{j}, j = 1,\dots,n$.

As an example, consider an USDMNN with three hidden layers. For fixed windows $W_{1}, W_{2}, W_{3} \in \mathcal{P}(E)$, this sequential architecture realizes the operators in $\Psi_{W}$, with $W = W_{1} \oplus W_{2} \oplus W_{3}$, in which $\oplus$ stands for the \textit{Minkowski addition}, that can be written as the composition of operators in $\Psi_{W_{1}}, \Psi_{W_{2}}$ and $\Psi_{W_{3}}$, that is, $\mathcal{H}(\mathcal{A}) = \{\psi \in \Psi_{W}: \psi = \psi^{W_{3}}\psi^{W_{2}}\psi^{W_{1}}; \psi^{W_{i}} \in \Psi_{W_{i}}, i = 1,2,3\}$.	For a proof of this fact, see \cite{barrera1996set}.

\subsection{Representation of USDMNN}
\label{Sec33}

The USDMNN realized by MCG $\mathcal{G} = (\mathcal{V},\mathcal{E},\mathcal{C})$ can be represented by a sequence $\{\psi_{1},\dots,\psi_{n}\}$, $n \geq 2,$ of $W$-operators with windows $W_{1},\dots,W_{n}$ as the composition
\begin{linenomath}
	\begin{equation}
		\label{def_MLWO}
		\psi_{\mathcal{G}} = \psi_{n} \circ \cdots \circ \psi_{1}
	\end{equation}
\end{linenomath}
in which $\psi_{j} = \mathcal{C}(\mathfrak{v}_{j}), j = 1,\dots,n$. Based on \eqref{def_MLWO}, we propose a representation for the class of operators realized by an USDMNN based on the window and characteristic function of the operators $\psi_{j}$.

For each $i= 1,\dots,n$, we fix a $d_{i} \geq 3$ odd and assume that each window $W_{i}, i = 1,\dots,n,$ is a connected subset of $F_{d_{i}} = \{- (d_{i} - 1)/2,\dots,(d_{i} - 1)/2\}^{2}$, the square of side $d_{i}$ centered at the origin of $E$. This means that, for every $w,w^{\prime} \in W_{i}$, there exists a sequence $w_{0},\dots,w_{r} \in W_{i}, r \geq 1,$ such that $w_{0} = w, w_{r} = w^{\prime}$ and $\lVert w_{i} - w_{i+1} \rVert_{\infty} = 1$, for all $i = 0,\dots,r-1$. Denoting $\mathscr{C}_{d} = \left\{W \subseteq F_{d}: W \text{ is connected}\right\}$, we assume that $W_{i} \in \mathscr{C}_{d_{i}}$ for all $i = 1,\dots,n$. 

For each $W \in \mathscr{C}_{d_{i}}$, let $\mathscr{B}_{W} = \{f: \mathcal{P}(W) \mapsto \{0,1\}\}$ be the set of all binary functions on $\mathcal{P}(W)$, and define $\mathscr{F}_{i} = \{(W,f): W \in \mathscr{C}_{d_{i}}, f \in \mathscr{B}_{W}\}$ as the collection of $W$-operators with window $W$ in $\mathscr{C}_{d_{i}}$ and characteristic function $f$ in $\mathcal{B}_{W}$, which are completely defined by a pair $(W,f)$. Finally, let $\Theta = \prod_{i=1}^{n} \mathscr{F}_{i}$ be the Cartesian product of $\mathscr{F}_{i}$. Observe that an element $\theta$ in $\Theta$ is actually a sequence of $n$ $W$-operators with windows in $\mathscr{C}_{d_{i}}, i = 1,\dots,n$, which we denote by $\theta = \{(W_{1},f_{1}),\dots,(W_{n},f_{n})\}$. Denoting the $W$-operator represented by $(W_{i},f_{i})$ as $\psi_{i}$, a $\theta \in \Theta$ generates a USDMNN $\psi_{\theta}$ via expression \eqref{def_MLWO}. 

The USDMNN architecture $\mathcal{A} = (\mathcal{V},\mathcal{E},\mathcal{F})$ with $n$ hidden layers and $\mathcal{F} = \{\iota\} \times \Psi_{F_{d_{1}}} \times \cdots \times \Psi_{F_{d_{n}}} \times \{\iota\}$ is such that $\mathcal{H}(\mathcal{A}) = \left\{\psi_{\theta}: \theta \in \Theta\right\}$, so $\Theta$ is a representation for the class of operators generated by the architecture $\mathcal{A}$. 

Since an operator locally defined in $W$ is also locally defined in any $W^{\prime} \supset W$ (cf. Proposition 5.1 in \cite{barrera1996set}) it follows that $\Theta$ is actually an overparametrization of $\mathcal{H}(\mathcal{A})$ since there are many representations $(W^{\prime},f_{\psi})$ for a same W-operator locally defined in $W \subsetneq F_{d_{i}}$. This overparametrization, discussed more generally in \cite{marcondes2023paradigm}, also happens in the representation of CDMNN proposed in \cite{marcondes2023discrete} and we take advantage of it to propose a SLDA to train USDMNN and mitigate the risk of overfitting.

\section{Training USDMNN via the stochastic lattice descent algorithm}
\label{SecTrain}

The training of a USDMNN is performed by obtaining a sequence $\hat{\theta} \in \Theta$ of $W$-operators via the minimization of an empirical error on a sample $\{(X_{1},Y_{1}),\dots,\\(X_{N},Y_{N})\}$, of $N$ input images $X$ and output target transformed images $Y$. In order to mitigate overfitting the sample, the training of a USDMNN will be performed by a two-step algorithm that, for a fixed sequence of windows, learns a sequence of characteristic functions by minimizing the empirical error $L_{t}$ in a training sample, and then learns a sequence of windows by minimizing an empirical error $L_{v}$ in a validation sample over the sequences of windows. More details about the empirical errors $L_{t}$ and $L_{v}$ will be given in Section \ref{SecToyExample}.

These two steps are instances of an algorithm for the minimization of a function in a subset of a Boolean lattice. On the one hand, the set $\mathscr{C} \coloneqq \prod_{i=1}^{n} \mathscr{C}_{d_{i}}$ of all sequences of $n$ connected windows is a subset of a Boolean lattice isomorphic to $\prod_{i=1}^{n} \{0,1\}^{F_{d_{i}}}$, so minimizing the validation error over the windows means minimizing a function in a subset of a Boolean lattice. On the other hand, for a fixed sequence of windows $(W_{1},\dots,W_{n}) \in \mathscr{C}$, the set $\mathscr{B}_{W_{1}} \times \cdots \times \mathscr{B}_{W_{n}}$ of all sequences of characteristic functions with these windows is a Boolean lattice isomorphic to $\{0,1\}^{\mathcal{P}(W_{1})} \times \cdots \times \{0,1\}^{\mathcal{P}(W_{n})}$, so minimizing the training error in this space means minimizing a function in a Boolean lattice. 

\begin{figure}[ht]
	\centering
	\includegraphics[width=\linewidth]{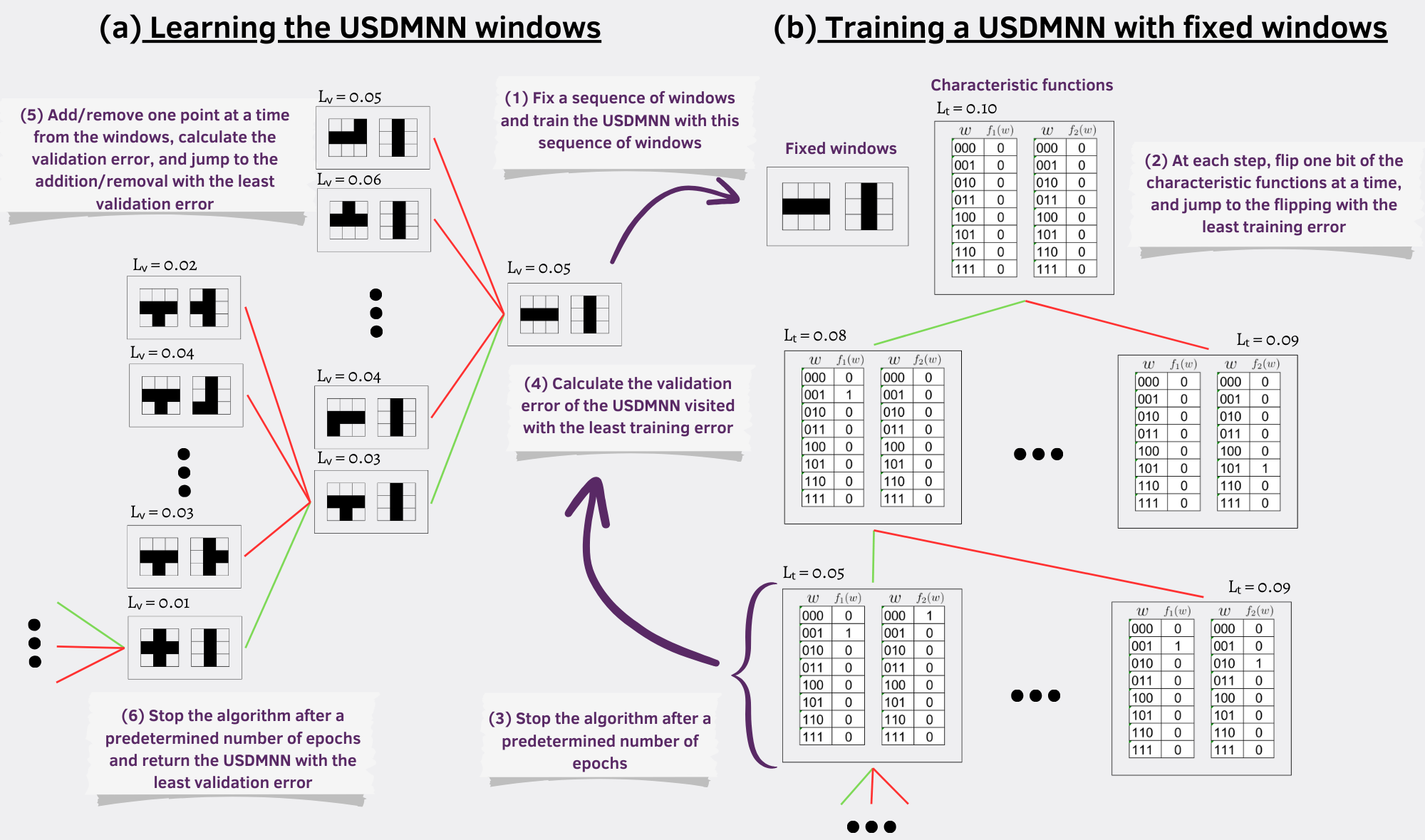}
	\caption{Illustration of the deterministic version of the SLDA to (a) learn the USDMNN windows and (b) train a USDMNN with fixed windows. To simplify the illustration, we considered only the possibility of adding a point to a window, and flipping a bit from 0 to 1 of a characteristic function, at each step, even though a point can be erased from a window, and a flipping from 1 to 0 may happen, if they have the least respective error. The training error $L_{t}$ or validation error $L_{v}$ of each point is on top of it.} \label{fig_alg}
\end{figure}

The U-curve algorithms \cite{u-curve3,ucurveParallel,reis2018,u-curve1} were proposed to minimize a U-curve function on a Boolean lattice. In summary, these algorithms perform a greedy search of a Boolean lattice, at each step jumping to the point at distance one with the least value of the function, and stopping when all neighbor points have a function value greater or equal to that of the current point. This greedy search of a lattice, that at each step goes to the direction that minimizes the function, is analogous to the dynamic of the gradient descent algorithm to minimize a function with domain in $\mathbb{R}^{p}$. Inspired by the U-curve algorithms and by the success of stochastic gradient descent algorithms for minimizing overparametrized functions in $\mathbb{R}^{p}$, and following the ideas of \cite{marcondes2023discrete}, we propose a stochastic lattice descent algorithm (SLDA) to train USDMNN. In Figure \ref{fig_alg}, we present the main ideas of the algorithm, and in Sections \ref{SecT1} and \ref{SecT2} we formally define it.

\subsection{Training a USDMNN with fixed windows}
\label{SecT1}

For each $\boldsymbol{W} \coloneqq \{W_{1},\dots,W_{n}\} \in \mathscr{C}$ let $\Theta_{\boldsymbol{W}} \coloneqq \{\{(W_{1},f_{1}),\dots,(W_{n},f_{n})\}: f_{i} \in \mathscr{F}_{W_{i}}, i = 1,\dots,n\}$ be all sequences of $W$-operators with windows $\boldsymbol{W}$. Observe there is a bijection between $\Theta_{\boldsymbol{W}}$ and the Boolean lattice $\prod_{i=1}^{n} \{0,1\}^{\mathcal{P}(W_{i})}$, and consider the Boolean lattice $(\Theta_{\boldsymbol{W}},\leq)$. 

Denoting by $d$ the distance in the acyclic directed graph $(\Theta_{\boldsymbol{W}},\leq)$, we define the neighborhood of a $\theta \in \Theta_{\boldsymbol{W}}$ as $N(\theta) = \left\{\theta^\prime \in \Theta_{\boldsymbol{W}}: d(\theta,\theta^\prime) = 1\right\}$. Observe that $\theta,\theta^{\prime} \in \Theta_{\boldsymbol{W}}$ are such that $d(\theta,\theta^\prime) = 1$ if, and only if, all their characteristic functions but one are equal, and in the one in which they differ, the difference is in only one point of their domain. In other words, $\theta^{\prime}$ is obtained from $\theta$ by flipping the image of one point of one characteristic function from 0 to 1 or from 1 to 0.

The SLDA for learning the characteristic functions of a USDMNN with fixed windows is formalized in Algorithm \ref{A2}. The initial point $\theta \in \Theta_{\boldsymbol{W}}$, a batch size $b$, the number $n$ of neighbors to be sampled at each step, and the number of training epochs are fixed. For each epoch, the training sample is randomly partitioned in $N/b$ batches, and we denote the training error on batch $j$ by $L_{t}^{(j)}$. For each batch, $n$ neighbors of $\theta$ are sampled and $\theta$ is updated to the sampled neighbor with the least training error $L_{t}^{(j)}$, that is calculated on the sample batch $j$. Observe that $\theta$ is updated at each batch, so during an epoch, it is updated $N/b$ times. At the end of each epoch, the training error $L_{t}(\theta)$ of $\theta$ on the whole training sample is compared with the error of the point with the least training error visited so far at the end of an epoch, and it is stored as this point if its training error is lesser. After the predetermined number of epochs, the algorithm returns the point with the least training error on the whole sample visited at the end of an epoch.

Observe that Algorithm \ref{A2} has two sources of stochasticity: the sampling of neighbors and the sample batches. If $n = n(\theta) \coloneqq |N(\theta)|$ and $b = N$, then this algorithm reduces to the deterministic one illustrated in Figure \ref{fig_alg}. Furthermore, the complexity of the algorithm is controlled by the number $n$ of sampled neighbors, the batch size $b$ and the number of epochs. See \cite{marcondes2023discrete} for a further discussion about a SLDA. 

\begin{algorithm}[ht]
	\centering
	\caption{Stochastic lattice descent algorithm for learning the characteristic functions of a USDMNN with fixed windows $\boldsymbol{W} = \{W_{1},\dots,W_{n}\}$.}
	\label{A2}
	\begin{algorithmic}[1]
		\ENSURE $\theta \in \Theta_{\boldsymbol{W}}, n, b, Epochs$		
		\STATE $L_{min} \gets L_{t}(\theta)$
		\STATE $\theta^{\mathbb{A}}_{\boldsymbol{W}} \gets \theta$ 
		\FOR{run $\in \{1,\dots,\text{Epochs}\}$}
		\STATE $\text{ShuffleBatches}(b)$
		\FOR{$j \in \{1,\dots,N/b\}$}
		\STATE $\tilde{N}(\theta) \gets \text{SampleNeighbors}(\theta,n)$	
		\STATE $\theta \gets \theta^{\prime} \text{ s.t. } \ \theta^{\prime} \in \tilde{N}(\theta) \text{ and } L_{t}^{(j)}(\theta^\prime)  = \min\{L_{t}^{(j)}(\theta^{\prime\prime}): \theta^{\prime\prime} \in \tilde{N}(\theta)\}$
		\ENDFOR
		\IF{$L_{t}(\theta) < L_{min}$}
		\STATE $L_{min} \gets L_{t}(\theta)$
		\STATE $\theta^{\mathbb{A}}_{\boldsymbol{W}} \gets \theta$ 
		\ENDIF
		\ENDFOR
		\RETURN{$\theta^{\mathbb{A}}_{\boldsymbol{W}}$}
	\end{algorithmic}
\end{algorithm}

\subsection{Learning the windows of USDMNN windows}
\label{SecT2}

In order to learn the windows of a USDMNN, we apply the SLDA on $\mathscr{C}$, that is a subset of the Boolean lattice $\prod_{i=1}^{n} \{0,1\}^{F_{d_{i}}}$. For each $\boldsymbol{W} \in \mathscr{C}$, let $L_{v}(\boldsymbol{W}) \coloneqq L_{v}(\theta^{\mathbb{A}}_{\boldsymbol{W}})$ be the validation error of the USDMNN realized by $\theta^{\mathbb{A}}_{\boldsymbol{W}}$, which was learned by Algorithm \ref{A2}. 

We define the neighborhood of $\boldsymbol{W}$ in $\mathscr{C}$ as $N(\boldsymbol{W}) = \left\{\boldsymbol{W}^\prime \in \mathscr{C}, d(\boldsymbol{W},\boldsymbol{W}^\prime) = 1\right\}$ in which $d$ means the distance in the acyclic directed graph $(\mathscr{C},\leq)$. Observe that $\boldsymbol{W},\boldsymbol{W}^\prime \in \mathscr{C}$ are such that $d(\boldsymbol{W},\boldsymbol{W}^\prime) = 1$ if, and only if, all their windows but one are equal, and in the one in which they differ, the difference is of one point. In other words, $\boldsymbol{W}^\prime$ is obtained from $\boldsymbol{W}$ by adding or removing one point from one of its windows.

The SLDA for learning the windows of a USDMNN is formalized in Algorithm \ref{A3}. The initial point $\boldsymbol{W} \in \mathscr{C}$, a batch size $b$, the number $n$ of neighbors to be sampled at each step, and the number of training epochs are fixed. For each epoch, the validation sample is randomly partitioned in $N/b$ batches, and we denote the validation error of any $\boldsymbol{W}$ on batch $j$ by $L_{v}^{(j)}(\boldsymbol{W}) \coloneqq L_{v}^{(j)}(\theta^{\mathbb{A}}_{\boldsymbol{W}})$ which is the empirical error on the $j$-th batch of the validation sample of the USDMNN realized by $\theta^{\mathbb{A}}_{\boldsymbol{W}}$, learned by Algorithm \ref{A2}. 

For each batch, $n$ neighbors of $\boldsymbol{W}$ are sampled and $\boldsymbol{W}$ is updated to the sampled neighbor with the least validation error $L_{v}^{(j)}$. At the end of each epoch, the validation error $L_{v}(\boldsymbol{W})$ of $\boldsymbol{W}$ on the whole validation sample is compared with the error of the point with the least validation error visited so far at the end of an epoch, and it is stored as this point if its validation error is lesser. After the predetermined number of epochs, the algorithm returns the point with the least validation error on the whole sample visited at the end of an epoch.

Algorithms \ref{A2} and \ref{A3} are analogous and differ only on the lattice ($\Theta_{\boldsymbol{W}}$ and $ \mathscr{C}$) they search and the function they seek to minimize (the training and the validation error).

\begin{algorithm}[ht]
	\centering
	\caption{Stochastic lattice descent algorithm for learning the windows of a USDMNN.}
	\label{A3}
	\begin{algorithmic}[1]
		\ENSURE $\boldsymbol{W} \in \mathscr{C}, n, b, Epochs$		
		\STATE $L_{min} \gets L_{v}(\boldsymbol{W})$
		\STATE $\boldsymbol{W}^{\mathbb{A}} \gets \boldsymbol{W}$ 
		\FOR{run $\in \{1,\dots,\text{Epochs}\}$}
		\STATE $\text{ShuffleBatches}(b)$
		\FOR{$j \in \{1,\dots,N/b\}$}
		\STATE $\tilde{N}(\boldsymbol{W}) \gets \text{SampleNeighbors}(\boldsymbol{W},n)$	
		\STATE $\boldsymbol{W} \gets \boldsymbol{W}^{\prime} \text{ s.t. } \ \boldsymbol{W}^{\prime} \in \tilde{N}(\boldsymbol{W}) \text{ and } L_{v}^{(j)}(\boldsymbol{W}^\prime)  = \min\{L_{v}^{(j)}(\boldsymbol{W}^{\prime\prime}): \boldsymbol{W}^{\prime\prime} \in \tilde{N}(\boldsymbol{W})\}$
		\ENDFOR
		\IF{$L_{v}(\boldsymbol{W}) < L_{min}$}
		\STATE $L_{min} \gets L_{v}(\boldsymbol{W})$
		\STATE $\boldsymbol{W}^{\mathbb{A}} \gets \boldsymbol{W}$ 
		\ENDIF
		\ENDFOR
		\RETURN{$\boldsymbol{W}^{\mathbb{A}}$}
	\end{algorithmic}
\end{algorithm}

\section{Application: Boundary recognition of digits with noise}
\label{SecToyExample}

As an example, we treat the problem of boundary recognition of digits with noise. We consider the USDMNN with two layers, and windows contained in the $3 \times 3$ square trained with a training sample of $10$, and a validation sample of $10$, $56 \times 56$ binary images. The training and validation samples are the same used in \cite{marcondes2023discrete} and the initial windows were considered as the five point cross centered at the origin. The initial characteristic functions for the first sequence of windows were randomly chosen, while the initial characteristic functions of a later sequence of windows were those that minimized the training error of its neighbor visited before, except for the window on which they differ, where the characteristic function is initiated randomly. The algorithms were implemented in \textbf{python} and are available at \url{https://github.com/MarianaFeldman/USDMM}. The training was performed on a personal computer with processor Intel Core i7-1355U x 12 and 16 GB of RAM.

We consider the intersection of union (IoU) error that is more suitable for form or object detection tasks. The IoU error of $\psi_{\theta}$ in the training sample is
\begin{linenomath}
	\begin{align}
		L_{t}(\theta) = 1 - \frac{1}{10} \sum_{k=1}^{10} \frac{|Y_{k} \cap \psi_{\theta}(X_{k})|}{|Y_{k} \cup \psi_{\theta}(X_{k})|} 
	\end{align}
\end{linenomath}
that is, the mean proportion of pixels in $Y \cup \psi_{\theta}(X)$ that are not in $Y \cap \psi_{\theta}(X)$ among the sample points in the training sample. The respective error on the validation sample is denoted by $L_{v}(\theta)$.

\begin{figure}[ht]
	\centering
	\includegraphics[width=\linewidth]{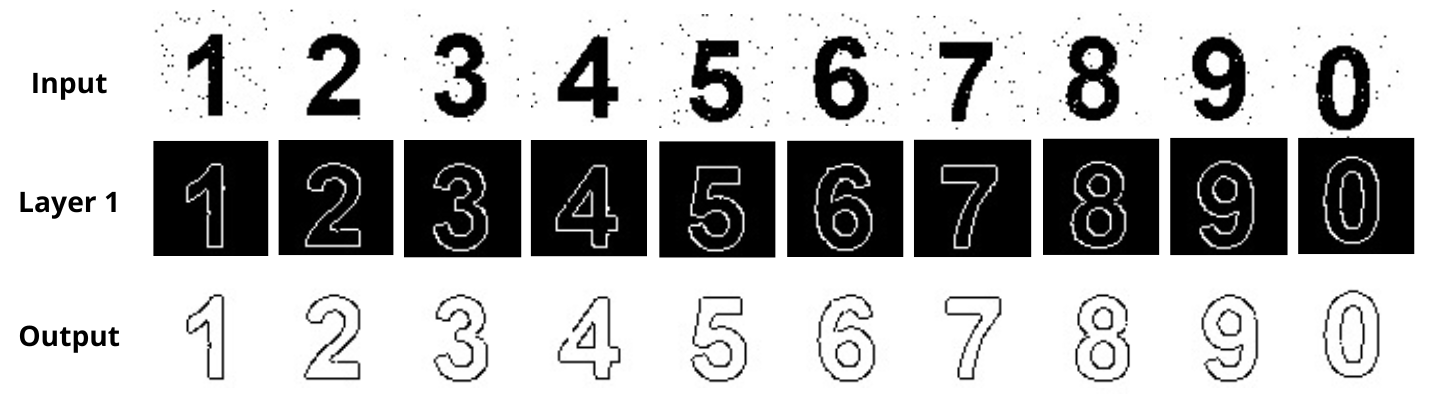}
	\caption{Result obtained after applying, to the validation sample, each layer of the USDMNN trained with a batch size of $10$ that had the least validation error.} \label{fig_exe}
\end{figure}

The results are presented in Table \ref{table_res} and Figure \ref{fig_exe}. We trained the two layer USDMNN with batch sizes $1$, $5$ and $10$, sampling $8$ neighbors in the SLDA for the characteristic functions and considering all neighbors in the SLDA for the windows. These batch sizes refer to the SLDA for the characteristic functions, and in all cases we considered a batch size of $10$ in the SLDA for the windows. We considered $50$ epochs to train the windows and $100$ epochs to train the characteristic functions. Each scenario was trained ten times, starting from distinct initial values of the characteristic functions. We present the minimum, mean and standard deviation of the results over the ten repetitions in Table \ref{table_res}.

The best results were obtained with batch size $10$, and it seems that, with the initial windows and characteristic functions we are considering, it is not possible to properly train with batch sizes $1$ and $5$. The training and validation error of the USDMNN trained with batch size $10$ did not vary a lot over the repetitions, so the algorithm is not sensible to the initial value for batch size $10$. Moreover, with batch size $10$, the SLDA for the characteristic functions took in average around $80$ epochs to reach the minimum in all repetitions, and the SLDA for the windows took an average of $26$ epochs to the minimum over the ten repetitions, although the variation was great and there was a repetition in which the minimum was achieved after only five epochs.

The time it took to train the USDMNN was greater than that it took to train CDMNN in \cite{marcondes2023discrete}, but it can be decreased with a more sophisticated implementation of the algorithms. The minimum training and validation error of the trained operator attained with USDMNN ($0.032$ and $0.052$) was slightly greater than the respective minimum error obtained in \cite{marcondes2023discrete} (around $0.029$ and $0.042$). Therefore, the USDMNN have obtained similar empirical results as the CDMNN in \cite{marcondes2023discrete} without strong prior information.

\begin{table}[ht]
	\centering
	\caption{\footnotesize Results of the two layer USDMNN trained with batch sizes $b = 1$, $b = 5$, and $b = 10$ for the characteristic function SLDA. The results are in the form Minimum - Average (Standard Deviation) over the ten repetitions. We present the minimum training error observed during training; the training and validation error of the trained USDMNN; the algorithm total time and time until the minimum validation error; the number of epochs to the minimum validation error; and the average number of epochs until the minimum training error with fixed windows.} \label{table_res}
	\resizebox{\linewidth}{!}{\begin{tabular}{|c|c|c|c|c|c|c|c|}
			\hline
			b & Min. Train & Learned Train & Min Val. & Total time (h) & Time to min. (h) & Epo. to min. (W) & Mean Epo. to min (f) \\ 
			\hline
			1 & 0.442 - 0.57 (0.07) & 0.597 - 0.677 (0.053) & 0.183 - 0.238 (0.031) & 113.4 - 135.6 (16.1) & 11.6 - 45.7 (37.2) & 4 - 16.8 (15.4) & 47.8 - 49.0 (0.8) \\ 
			5 & 0.421 - 0.49 (0.04) & 0.492 - 0.611 (0.076) & 0.127 - 0.155 (0.015) & 78.5 - 91.9 (20.3) & 13.648 - 32.478 (19.0) & 4 - 18.7 (11.9) & 51.4 - 52.9 (1.3) \\ 
			10 & 0.031 - 0.036 (0.004) & 0.032 - 0.039 (0.006) & 0.052 - 0.056 (0.003) & 173.6 - 182.6 (13.3) & 14.262 - 98.324 (70.9) & 5 - 26.2 (17.7) & 80.9 - 82.6 (1.3) \\
			\hline
	\end{tabular}}
\end{table}

\section{Next steps and future research}
\label{SecPerspectives}

In this paper, we proposed a hierarchical SLDA to train USDMNN and illustrated it in a simple example. We obtained empirical results almost as good as that of \cite{marcondes2023discrete}, but without the necessity to design a DMNN based on prior information. The next step of this work is to develop an efficient implementation of the algorithms and perform a more extensive experimental study to better understand features of them. In particular, it is necessary to better understand the sensitivity to the initial values and the role of the batch size on both the SLDA for the windows and for the characteristic functions. A more thorough comparison with the CDMNN proposed in \cite{marcondes2023discrete} is also necessary.

We are currently working on an efficient implementation of the hierarchical SLDA that, we believe, will significantly decrease the training time, so methods based on USDMNN may also be competitive with CDMNN from a computational complexity perspective. With a more efficient algorithm, it will be possible to fine train USDMNN and better study the effect of settings such as the number of layers and the window size on the performance. An efficient method to train DMNN without the need of strong prior information may help popularize methods based on them among practitioners which do not have strong knowledge in mathematical morphology to design specific CDMNN architectures.

A promising line of research is to compare the USDMNN and CDMNN with other methods proposed in the literature. Since the DMNN are exact representations of morphological operators, an interesting study would be to compare them with other exact methods, rather than with approximations such as the classical MNN. In particular, it would be interesting to compare the DMNN with the binary morphological neural networks proposed in \cite{aouad2022binary,aouad2023foundation}. We leave such a comparison as a topic for future research.

%
%
%
\bibliographystyle{splncs04}
\bibliography{ref}
\end{document}